\documentclass[10pt,twocolumn,letterpaper]{article}

\usepackage{wacv}

\makeatletter
\@namedef{ver@everyshi.sty}{}
\makeatother

\usepackage{times}
\usepackage{epsfig}
\usepackage{graphicx}
\usepackage{amsmath}
\usepackage{amssymb}

\usepackage{nicefrac}
\usepackage{subcaption}   

\usepackage{tikz}
\usetikzlibrary{decorations.pathreplacing,shapes,positioning,arrows}
\usepackage{pgfplots}

\usepackage{caption,booktabs,array} 

\usepackage{footmisc}   

\usepackage{url}

\usepackage{times}
\usepackage{epsfig}
\usepackage{graphicx}
\usepackage{amsmath}
\usepackage{amssymb}

\usepackage{textcomp}
\usepackage{subcaption}
\usepackage{booktabs}
\usepackage[nocompress]{cite}  
\usepackage{tablefootnote}

\definecolor{BrightRoyalPurple}{rgb}{0.65, 0.05, 0.78}
\definecolor{DarkGreen}{rgb}{0.0, 0.4, 0.0}

\definecolor{Green}{rgb}{0.0, 0.7, 0.0}

\definecolor{Blue}{rgb}{0.0, 0.0, 1.0}

\newcommand{\comment}[1]{}

\graphicspath{{figures/}}

\wacvfinalcopy 

\def\wacvPaperID{562} 

\ifwacvfinal\fi
\begin{document}

\title{Smart Hypothesis Generation for Efficient and Robust Room Layout Estimation}

\author{Martin Hirzer$^1$ \\
{\tt\small hirzer@icg.tugraz.at}
\and
Peter M. Roth$^1$ \\
{\tt\small pmroth@icg.tugraz.at}
\and
Vincent Lepetit$^{2,1}$ \\
{\tt\small vincent.lepetit@u-bordeaux.fr} \\
\and
\small $^1$Institute of Computer Graphics and Vision, Graz University of Technology, Austria\\
\and
\small $^2$Laboratoire Bordelais de Recherche en Informatique, University of Bordeaux, France\\
}

\maketitle
\ifwacvfinal\thispagestyle{empty}\fi

We propose a novel method to efficiently estimate the spatial layout of a room from a single monocular RGB image. As existing approaches based on low-level feature extraction, followed by a vanishing point estimation are very slow and often unreliable in realistic scenarios, we build on semantic segmentation of the input image. To obtain better segmentations, we introduce a robust, accurate and very efficient hypothesize-and-test scheme. The key idea is to use three segmentation hypotheses, each based on a different number of visible walls. For each hypothesis, we predict the image locations of the room corners and select the hypothesis for which the layout estimated from the room corners is consistent with the segmentation. We demonstrate the efficiency and robustness of our method on three challenging benchmark datasets, where we significantly outperform the state-of-the-art.

\section{Introduction}
\label{sec:intro}

Room layout estimation from a monocular RGB image aims at finding the boundaries of the floor, ceiling, and the individual walls in an image, as depicted in Fig.~\ref{fig:teaser}. Identifying these semantically important regions is beneficial for a wide range of applications, including indoor navigation, object detection, scene reconstruction, and augmented reality. For these applications, it would be highly relevant to know which features are related to the fixed background or to movable foreground objects (\eg, furniture) to guide robust object detection and recognition.

\newcommand{\figWidth}{0.95}
\begin{figure}[t]
  \centering\includegraphics[width=\figWidth\linewidth]{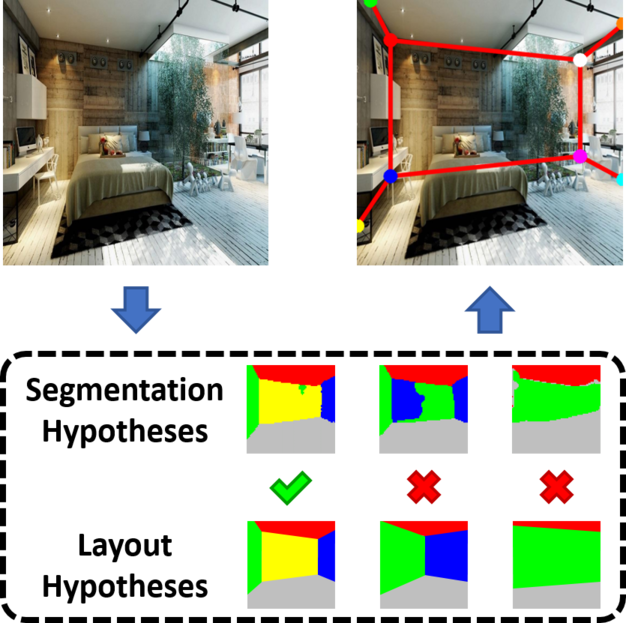}
  \caption{\label{fig:teaser} Estimating the room layout from a single given RGB image: We divide the task into three sub-problems, generate a segmentation and a layout hypothesis for each of them, and select the one that has the highest consistency.}
\end{figure}

However, the task is inherently challenging, since indoor scenes typically suffer from considerable amounts of clutter, varying lighting, and large intra-class variance. Moreover, the region boundaries that we are interested in are often severely occluded by furniture, preventing a direct inference. Hence, motivated by a large number of practical applications and still unresolved problems, there has been a considerable scientific interest within the last years. Most of these approaches are based on the extraction of low-level features followed by a ranking step in order to evaluate a potentially huge number of layout hypotheses, which is computationally expensive and severely limits their practical application~\cite{Hedau09, Ramalingam13, Lee09, Lee10, Schwing12, Wang10, Dasgupta16, Mallya15, Zhao17}. In contrast, \cite{Lee17} tries to overcome this drawback by directly predicting an ordered set of 2D keypoints, however, at the cost of requiring an additional, vulnerable room type classifier in order to correctly merge the keypoints into a layout.

To overcome these problems, we introduce an efficient and robust approach, where the key idea, as shown in Fig.~\ref{fig:teaser}, is to generate and evaluate three layout hypotheses (for one, two, or three visible walls). To this end, we first compute a segmentation based on each hypothesis and then predict the locations of the 2D keypoints defining the layout. Finally, we compare each layout generated from the 2D keypoints to its corresponding image segmentation and select the hypothesis that provides the best match.

This approach has several advantages: First, it allows us to automatically resolve the inherent ambiguity considering the left, center, and right wall regions of a room~\cite{Dasgupta16}.  
Second, in combination with  the derived semantics,  the wall-based hypotheses can be used to directly infer the layout from the keypoints. In particular, we do not rely on an additional room classification step or require the evaluation of a large set of layout hypotheses, which is an advantage over many previous works such as~\cite{Lee17, Hedau09, Ramalingam13, Mallya15, Dasgupta16}.
Third, using the semantic segmentation as an intermediate representation to predict the 2D keypoints improves the generalization capabilities, compared to predicting the 2D  keypoints directly from the image as in~\cite{Lee17}.

These benefits can also be seen from the experimental results, where we compare our approach to the state-of-the-art on three different publicly available benchmark datasets, namely the Large-scale Scene Understanding Challenge~(LSUN) room layout dataset~\cite{Zhang16}, the Hedau dataset~\cite{Hedau09}, and the NYUv2~303 dataset~\cite{Zhang13}. In fact, our method is not only very efficient, but also clearly outperforms existing approaches.

The remainder of the paper is organized as follows: First, in Section~\ref{sec:rw}, we discuss the related work on room layout estimation. Then, in Section~\ref{sec:method}, we introduce our new  approach based on smart semantic hypothesis generation. Next, in Section~\ref{sec:experiments}, we give a quantitative and qualitative comparison of our approach to the state-of-the-art and also provide an ablation study. Finally, in Section~\ref{sec:conclusion}, we summarize and conclude our work.

\section{Related Work}
\label{sec:rw}

We classify existing room layout estimation approaches into three main categories: (1) Bottom-up approaches, which first extract low-level features from the image and then generate and rank layout hypotheses based on vanishing points estimated from the aggregated features; (2) segmentation-based approaches, which follow a similar strategy but avoid the usage of hand-crafted features; (3) top-down approaches, which directly estimate an ordered set of 2D keypoints that define the layout.

\vspace {0.1cm}
\noindent 
\textbf{Bottom-Up}
One of the first bottom-up methods was Hedau~\etal~\cite{Hedau09}, which cluster line segments in order to detect three orthogonal vanishing points, generate layout candidates from the obtained points, and finally rank them using a structured SVM. Ramalingam~\etal~\cite{Ramalingam13} follow a similar approach but replace the line segments by line junctions. Lee~\etal~\cite{Lee09} introduce an orientation map based on line segments in order to reason about the layout. Schwing~\etal~\cite{Schwing12} try to speed up the structured layout prediction by transferring the concept of integral images~\cite{Viola04} to geometry. Wang~\etal~\cite{Wang10} use latent variables in order to jointly infer the layout and the clutter, and Lee~\etal~\cite{Lee10} incorporate object hypotheses to improve the final layout prediction. The main drawback of such methods, however, is that for many practical applications the required low-level features cannot be reliably estimated, making these methods prone to errors in realistic scenarios that contain lots of occlusions, clutter, and diverse lighting.


\noindent 
\textbf{Segmentation-based}
With the development of Deep Learning, there has been considerable interest to improve the low-level feature extraction by leveraging recent advances in semantic segmentation \cite{Mallya15,Dasgupta16,Ren16,Zhao17}. Building on fully convolutional networks~(FCNs)~\cite{Long15}, Mallya and Lazebnik~\cite{Mallya15}, Ren~\etal~\cite{Ren16}, and Zhao~\etal~\cite{Zhao17} estimate ``informative edge maps", whereas Dasgupta~\etal\cite{Dasgupta16} directly predict semantic surface labels (\ie, floor, ceiling, left, center, and right wall). The main differences between these approaches are amount and complexity of the required training data, ranging from simple box layouts typically available for the task at hand~\cite{Mallya15,Dasgupta16,Ren16} to very rich and detailed furniture segmentation masks that are hard to acquire~\cite{Zhao17}. Moreover, these methods still rely on vanishing point/line sampling followed by a layout generation and ranking step or require a computationally expensive optimization based on physical constraints in order to fit the final layout, which is cumbersome and slow.


\noindent 
\textbf{Top-Down}
Lee~\etal~\cite{Lee17}, on the other hand, follow a more direct, top-down approach. In particular, they try to directly estimate an ordered set of 2D keypoints that fully defines the layout. While this allows them to avoid the slow layout generation and ranking step, they require an explicit classification of the room type to infer the correct layout from the keypoints. However, given the inherent imbalance in the distribution of room types in typical indoor images, the accuracy of the classifier is rather low, specifically on the underrepresented types.



In this work, we also follow a top-down strategy by first estimating a set of ordered 2D keypoints, which can then be directly connected to generate the full layout. In contrast to~\cite{Lee17}, however, we avoid such an explicit room type classifier and instead exploit powerful semantic segmentation, which allows us to merge the obtained keypoints into a layout prediction much more conveniently. Specifically, we show that this can be achieved by evaluating only three layout hypotheses, which makes our approach also pretty fast.

Besides these main directions, there are also approaches that ease the problem by exploiting additional information such as depth~\cite{Zhang13}, floor plans~\cite{Liu15}, full 360\textdegree-panoramas~\cite{Zou18}, or by assuming that people are present in the scene in order to be able to reason about the layout~\cite{Chao13}. However, these requirements are often not fulfilled in realistic scenarios, which severely limits the practical applicability of these approaches.


\renewcommand{\figWidth}{1.0}
\begin{figure*}[!ht]
  \centering
  \includegraphics[width=\figWidth\linewidth]{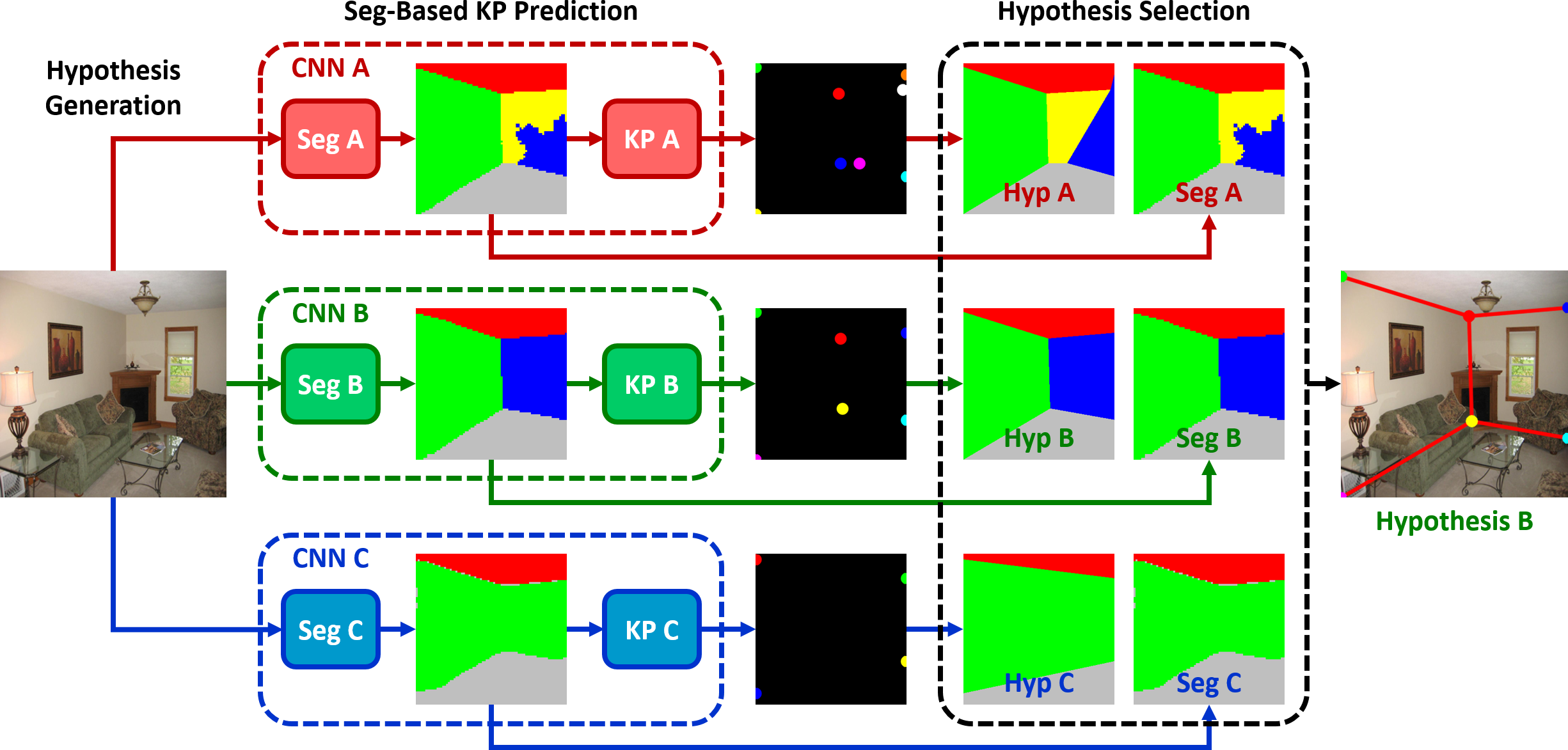}
  \caption{Overall system: Given an RGB input image, we first generate three hypotheses A, B, and C each assuming a different wall configuration, then predict the corresponding layouts, and finally select the one that best fits its associated segmentation. In the depicted example, hypothesis B provides the best fit, whereas A tries to fit a third wall that is not present, and C is too simplistic, not being able to explain the two wall configuration. Best viewed in color.}
  \label{fig:system}
\end{figure*}

\section{Smart Hypothesis Generation for Room Layout Estimation}
\label{sec:method}

In the following, we introduce our new robust room layout estimation approach, which is illustrated in Fig.~\ref{fig:system}.
Instead of forcing a  single model to handle all possible cases, we generate three layout hypotheses assuming that one, two, or three walls are visible in the image.
For each hypothesis, we first compute a semantic segmentation of the input image and estimate the 2D location of the layout's keypoints (Sections~\ref{sec:seg_based_kp} and~\ref{sec:hyp_gen}). To obtain the final layout, we then select the hypothesis that best fits its associated segmentation (Section~\ref{sec:hyp_sel}).

\subsection{Segmentation-based Keypoint Prediction}
\label{sec:seg_based_kp}

Directly predicting an ordered set of keypoints to estimate the layout of a room has proven to be superior to methods that first aggregate a set of low- or mid-level features and then generate and rank a multitude of layout hypotheses based on the gathered image cues~\cite{Lee17}. Thus, we follow this direction and design a network that takes an RGB image as input and outputs a set of ordered keypoint locations. In contrast to~\cite{Lee17}, however, we exploit a semantic segmentation as an intermediate network representation, which, in combination with the task-specific hypothesis generation described in Section~\ref{sec:hyp_gen}, enables us to avoid an explicit room type classifier.

\pagebreak

Specifically, similar to~\cite{Lee17}, we employ SegNet~\cite{Badrinarayanan15} as the base architecture of our network, since it is time and memory efficient and has shown good performance in various segmentation tasks. Like most semantic segmentation architectures, it consists of two sub-networks, an encoder and a decoder. The encoder applies a series of convolution and pooling operations, mapping the input image to lower resolution feature maps. The decoder then samples the low-resolution feature maps back up to the full image resolution for pixel-accurate classification. This is achieved by a series of non-linear upsampling operations based on the corresponding pooling indices in the encoder. Since the upsampled maps are sparse, they are convolved with learnable filters in order to produce dense feature maps.

The first part of our network is a standard SegNet, taking an RGB image of a size of $320 \times 320$ pixels as input and producing a semantic segmentation consisting of the following five classes: floor, ceiling, left, center, and right wall. However, we do not sample the low-resolution feature maps back up to the full image resolution, but cap the decoder at a size of $80 \times 80$ pixels, since we found this to be accurate enough in order to predict the keypoint locations. The second part of our network is a reduced version of SegNet, where both the encoder and decoder are capped at $80 \times 80$ pixels. It takes the output of the first part as input and predicts a set of ordered keypoint locations in the form of 2D Gaussian heatmaps~\cite{Lee17} of size $80 \times 80$ pixels.

\subsection{Wall-based Hypothesis Generation}
\label{sec:hyp_gen}

If we could always assume the same room type (\ie, a fixed keypoint configuration), predicting the 2D keypoints of a room via a semantic segmentation as an intermediate representation would be rather easy. However, in practice, the room type is not known in advance, making the problem more difficult.
%
%
Hence, Lee~\etal~\cite{Lee17} predict the 48 keypoint locations for all 11 room types defined in~\cite{Zhang16} simultaneously and then rely on an explicit type classifier attached to their network in order to identify the correct subset and order of keypoints. However, this approach is rather vulnerable, since its performance crucially depends on the accuracy of the classifier\footnote{Note that the classification accuracy reported in~\cite{Lee17} is only 81.5\%.}. This is particularly evident in images of less common room types, as we will show in Section~\ref{sec:experiments}.

In contrast, we propose a more robust, integrated solution, where we tackle the problem by generating three layout hypotheses based on the number of visible walls.
Thus, we start by first identifying three groups of rooms within the set of 11 types defined in~\cite{Zhang16} and shown in Fig.~\ref{fig:room_types}:

\begin{itemize}
  \vspace{-0.15cm}
  \item Group A: 3 visible walls (room types 0, 1, 2, and 7)
  \vspace{-0.15cm}
  \item Group B: 2 visible walls (room types 3, 4, 5, and 10)
  \vspace{-0.15cm}
  \item Group C: 1 visible wall (room types 6, 8, and 9)
\end{itemize}

\renewcommand{\figWidth}{0.086}
\begin{figure*}[ht]
  \centering
  \begin{subfigure}{\figWidth\linewidth}
    \centering
    \includegraphics[width=\linewidth]{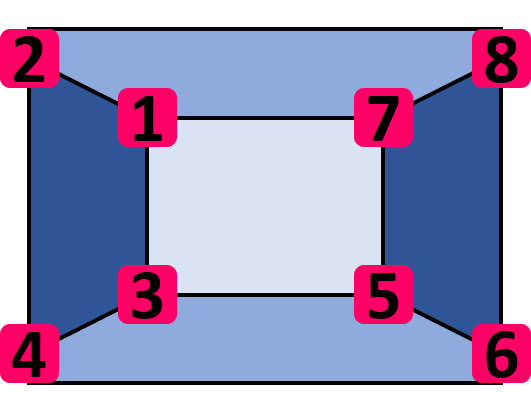}
    \caption*{\textcolor{red}{\underline{\textbf{Type 0}}}}
  \end{subfigure}
  \begin{subfigure}{\figWidth\linewidth}
    \centering
    \includegraphics[width=\linewidth]{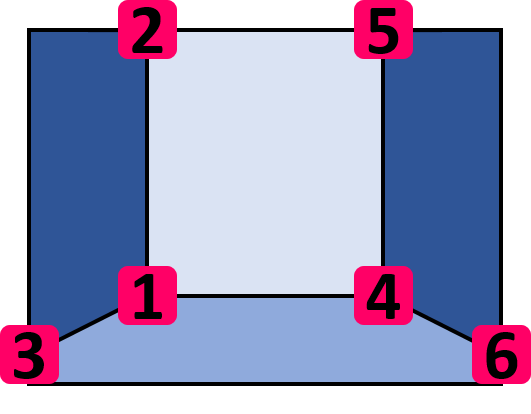}
    \caption*{\textcolor{red}{Type 1}}
  \end{subfigure}
  \begin{subfigure}{\figWidth\linewidth}
    \centering
    \includegraphics[width=\linewidth]{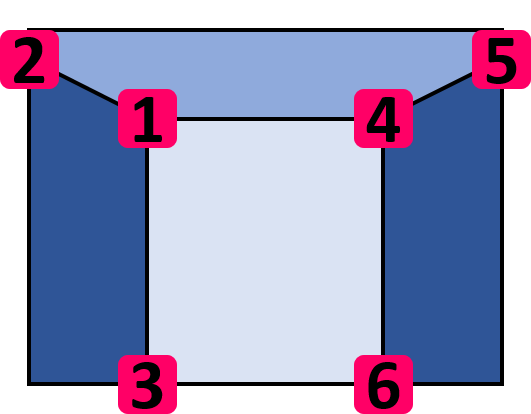}
    \caption*{\textcolor{red}{Type 2}}
  \end{subfigure}
  \begin{subfigure}{\figWidth\linewidth}
    \centering
    \includegraphics[width=\linewidth]{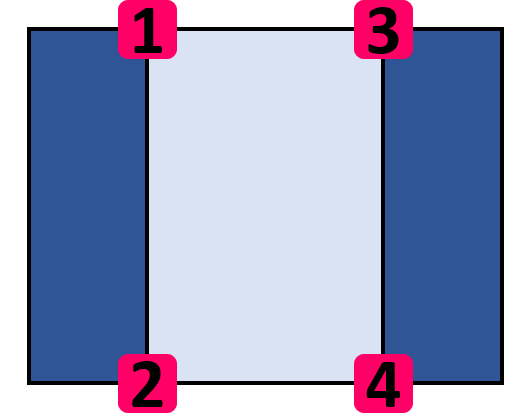}
    \caption*{\textcolor{red}{Type 7}}
  \end{subfigure}
  \begin{subfigure}{\figWidth\linewidth}
    \centering
    \includegraphics[width=\linewidth]{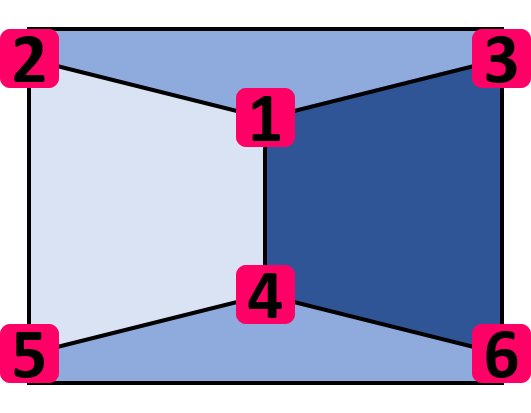}
    \caption*{\textcolor{green}{\underline{\textbf{Type 5}}}}
  \end{subfigure}
  \begin{subfigure}{\figWidth\linewidth}
    \centering
    \includegraphics[width=\linewidth]{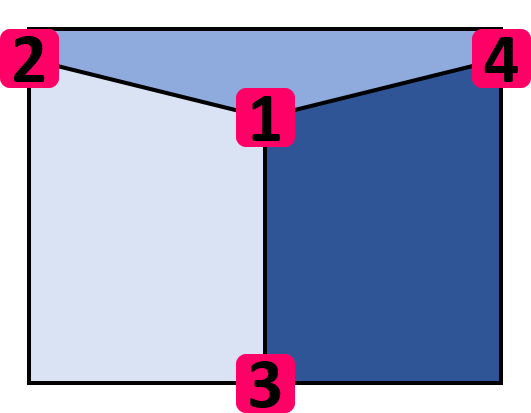}
    \caption*{\textcolor{green}{Type 3}}
  \end{subfigure}
  \begin{subfigure}{\figWidth\linewidth}
    \centering
    \includegraphics[width=\linewidth]{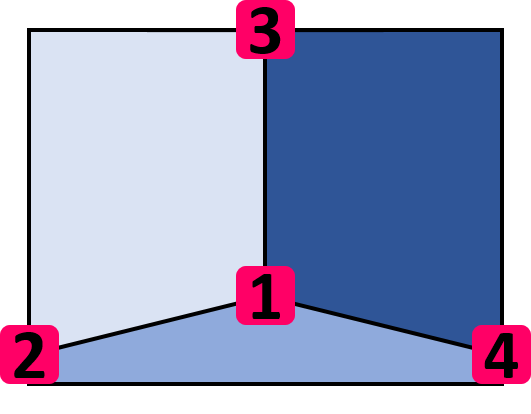}
    \caption*{\textcolor{green}{Type 4}}
  \end{subfigure}
  \begin{subfigure}{\figWidth\linewidth}
    \centering
    \includegraphics[width=\linewidth]{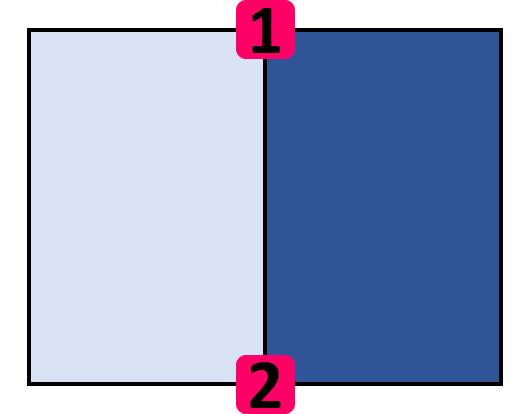}
    \caption*{\textcolor{green}{Type 10}}
  \end{subfigure}
  \begin{subfigure}{\figWidth\linewidth}
    \centering
    \includegraphics[width=\linewidth]{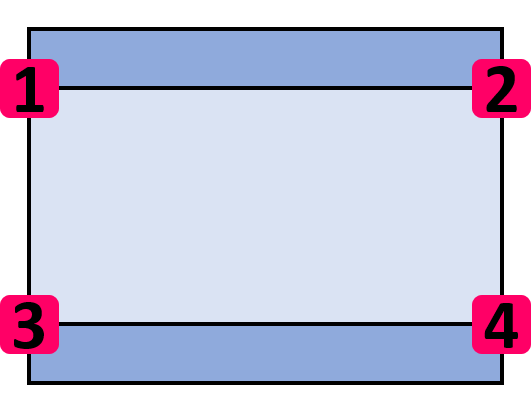}
    \caption*{\textcolor{blue}{\underline{\textbf{Type 6}}}}
  \end{subfigure}
  \begin{subfigure}{\figWidth\linewidth}
    \centering
    \includegraphics[width=\linewidth]{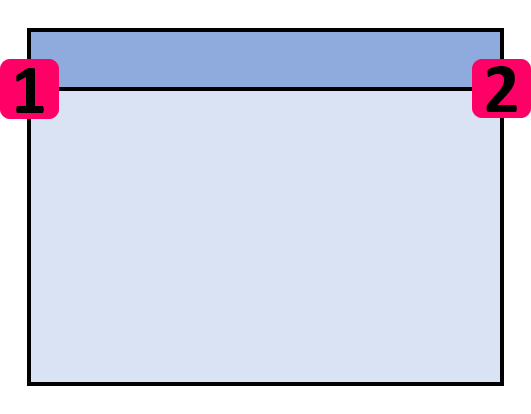}
    \caption*{\textcolor{blue}{Type 8}}
  \end{subfigure}
  \begin{subfigure}{\figWidth\linewidth}
    \centering
    \includegraphics[width=\linewidth]{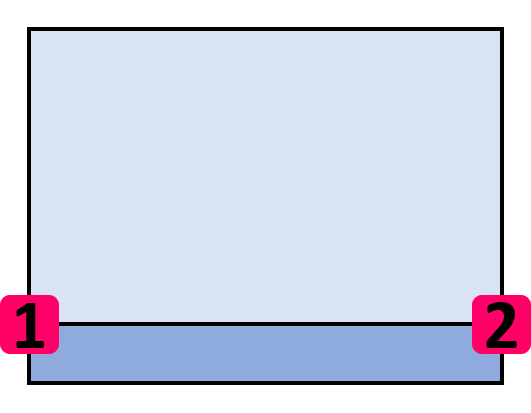}
    \caption*{\textcolor{blue}{Type 9}}
  \end{subfigure}
  \caption{The 11 room types with their respective keypoint order as defined in~\cite{Zhang16}. Note that rooms within the same group share parts of their configuration, only differing in the optional floor and ceiling regions. The configuration that contains both of the optional regions is considered the prototype of the corresponding group. Group A is marked in red, group B in green, and group C in blue, with the respective prototypes being highlighted. Best viewed in color.}
  \label{fig:room_types}
\end{figure*}

Rooms within one group share large parts of their layout configuration, except for the optional floor and ceiling region. This consistency can be exploited, not only to increase the accuracy and robustness of the segmentation and keypoint prediction, but also to infer the correct layout from the keypoints without requiring an auxiliary classification step. Additionally, it also allows us to implicitly handle the inherent ambiguity in the labels of the left, center, and right wall~\cite{Dasgupta16}. To the best of our knowledge, we are the first to take advantage of this consistency in the room layouts.

First, we re-arrange the keypoints defined in~\cite{Zhang16} to maximize the coherence within each group. For each group, we select the room type that contains all the keypoints, \ie, the type that contains both, floor and ceiling, as the prototype (see Fig.~\ref{fig:room_types}). Then, we re-arrange the keypoints of the other types to match the order of their respective prototype, as illustrated in Fig.~\ref{fig:re_order_kps} for type~4. Since keypoints belonging to the floor or ceiling are optional, the sequence of keypoint IDs is no longer required to be continuous.

\renewcommand{\figWidth}{0.75}
\begin{figure}[ht]
  \centering
  \includegraphics[width=\figWidth\linewidth]{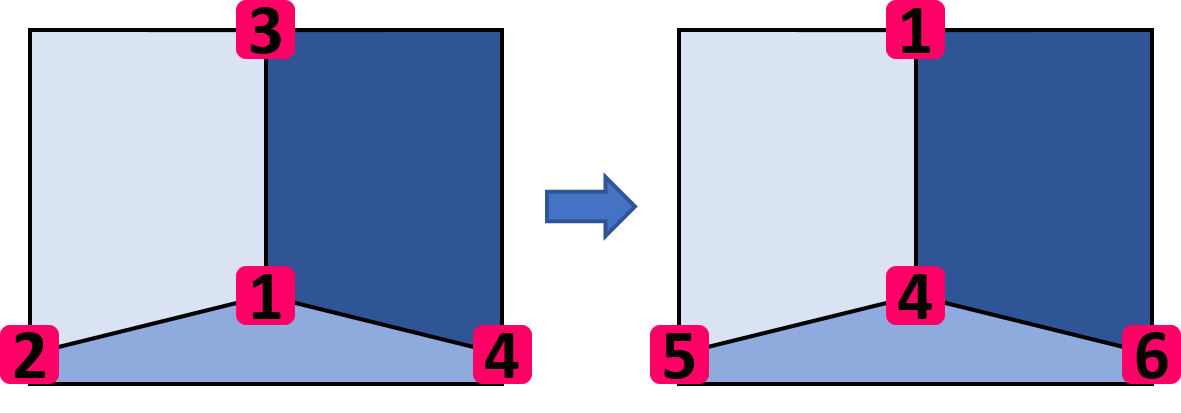}
  \caption{Keypoints of room type 4 are re-arranged to match the order of the corresponding prototype (type 5). Note that the keypoint IDs are not required to be continuous.}
  \label{fig:re_order_kps}
\end{figure}

Once the keypoints are re-arranged, we can use the same network architecture for all rooms within the same group, since they all share the same keypoint configuration. Thus, for each group, we train a separate CNN which tries to predict the keypoint locations of the respective room prototype, \ie, 8 keypoints according to type 0 for group A, 6 keypoints according to type 5 for group B, and 4 keypoints according to type 6 for group C. When inferring the layout from the keypoints, we have to decide whether to take the optional floor and ceiling keypoints into account. Conveniently, we can again exploit our obtained semantic segmentation for this task, by simply checking for a floor and ceiling region in the segmentation mask. This way, we can automatically derive the proper layout for each group, without an explicit classification step.

\subsection{Hypothesis Selection}
\label{sec:hyp_sel}

However, given an input image, we still have to decide which of the three groups to choose for the final layout. Although initial experiments indicated that directly classifying the layout group from the image works better than predicting the exact room type as in~\cite{Lee17} (which is no surprise since the task is easier), the performance was still not satisfactory. Thus, we introduce a more robust, integrated solution, where we forward the input image to all three groups simultaneously, generating three layout hypotheses. Then, in order to select the correct layout hypothesis, we can once more exploit our semantic segmentation. In particular, for each hypothesis, we compare the layout prediction to the semantic segmentation and pick the one that best fits its corresponding segmentation mask, as illustrated in Fig.~\ref{fig:system}. For evaluating the hypotheses, we define
\begin{equation}
  S_i = N_{mr}(L_i, S_i) + \lambda \cdot \text{mIoU}(L_i, S_i)
  \label{eq:hypo_score}
\end{equation}
as the matching score, where $i\in\{A, B, C\}$, $N_{mr}$ is the number of matching regions, $\text{mIoU}$ is the mean intersection over union~(IoU) over all regions between layout $L_i$ and segmentation $S_i$, and $\lambda$ is a weight term. The number of matching regions describes how many regions can be described by the layout, \ie, how many of the corresponding regions have an IoU greater than 80\%. Note that we do not normalize $N_{mr}$ by the overall number of regions since we want to put more emphasis on layouts that can ``explain'' many regions of their respective segmentation. Otherwise, simpler layouts would be preferred, as it is often easier to fit a single wall instead of two or three individual walls. In our case, setting $\lambda$ to 1 gave us the best results.

The key aspect is that each CNN is only trained on rooms from its specific group. Thus, both the segmentation and the keypoint prediction are very likely to fail if confronted with an image showing an unfamiliar type, which results in a low matching score. This can be seen very well from hypothesis A in Fig.~\ref{fig:system}. As a result, only the proper hypothesis achieves a high score and will be automatically selected. Moreover, by dividing the task into three sub-problems and tackling each of them with a specifically trained version of our CNN defined in Section~\ref{sec:seg_based_kp}, we can also appropriately handle the ambiguity typically encountered in the labels of the walls. This is in contrast to approaches that try to force a single CNN to handle all cases, which typically results in mixed up wall labels, as can be seen in~\cite{Dasgupta16} (similar to the center/right wall from hypothesis A in Fig.~\ref{fig:system}).


\section{Experiments}
\label{sec:experiments}

In this section, we evaluate our approach on three challenging room layout benchmarks, in particular the Large-scale Scene Understanding Challenge~(LSUN) room layout dataset~\cite{Zhang16}, the Hedau dataset~\cite{Hedau09}, and the NYUv2~303 dataset~\cite{Zhang13}. LSUN contains 4000 training, 394 validation, and 1000 test images that are sampled from the SUN database~\cite{Xiao10}. Hedau consists of 209 training, 53 validation, and 105 test images collected from the web and from LabelMe~\cite{Russell08}. NYUv2~303 is a randomly chosen subset of 202 training and 101 test images from the NYU-RGBD-v2 dataset~\cite{Silberman12}. All three benchmarks provide a diverse and challenging collection of indoor scenes containing clutter, occlusions, and varying lighting.

\subsection{Experimental Setup}
\label{sec:setup}

In our experiments, we follow the common practice of re-scaling all
input images to $320 \times 320$ pixels, and training our model on the
LSUN training set only~\cite{Lee17,Dasgupta16}. For testing, we run
our method and a re-implementation of RoomNet~(basic)~\cite{Lee17} as
a baseline on the corresponding test sets on the original image
scales, using the LSUN room layout challenge
toolkit~\cite{Zhang16}\footnote{As the code for \cite{Lee17} is not
  available, we re-implemented the method closely following the given implementation details. For LSUN, the ground truth for the test set is not available, so we evaluated on the validation split.}.

In order to evaluate the performance of our method, we use two standard metrics:
\begin{itemize}
  \item Pixel Error~(PE): pixelwise error between predicted surface labels and ground truth labels averaged over all images
  \item Keypoint Error~(KPE): Euclidean distance between predicted keypoints and ground truth positions, normalized by the image diagonal and averaged over all images
\end{itemize}

When training our three CNNs, we found that jointly training the segmentation and the keypoint prediction was difficult, in particular due to a more elaborate data augmentation used in the segmentation stage, which was not applicable to the keypoint predictor. Hence, we first trained the segmentation stage alone, followed by training the whole network while keeping the segmentation weights fixed. Naturally, the three networks A, B, and C were only trained on images corresponding to their respective group. However, for the segmentation part, learning turned out to be more stable by initializing the three specialized models with the weights of a general base model trained on all images.

For the segmentation part, we use the following training setup: stochastic gradient decent~(SGD), batch size~$14$, momentum~$0.99$, weight decay~$5e\mathrm{-}4$, and dropout rate~$0.5$. At the beginning, all weights are initialized using the method presented in~\cite{He15}. Furthermore, we apply batch normalization~\cite{Ioffe15} and the ReLU activation function~\cite{Nair10} after each convolution layer. The base model is trained for $200K$ iterations with an initial learning rate of $1e\mathrm{-}3$, which is reduced by factor of $5$ after $100K$ and $150K$ iterations, respectively. As expected, the resulting model has difficulties assigning the correct wall labels due to the inherent ambiguity, as has also been reported in~\cite{Dasgupta16}. For fine-tuning the specialized versions, we train each of them for another $100K$ iterations with an initial learning rate set to $1e\mathrm{-}4$, reduced by a factor of $5$ after $50K$ and $75K$ iterations. Note that for network~C, a slightly lower initial learning rate of $1e\mathrm{-}5$ is required, presumably caused by the rather limited amount of training images for that group. As data augmentation, we randomly apply horizontal mirroring, small variations in image lightness, and gentle affine transformations.

Training the keypoint prediction part seems to be easier, most likely due to the well-suited intermediate representation obtained via the semantic segmentation stage. Thus, we can directly learn each of the three keypoint predictors from scratch, without having to train a common base model for initialization first. The settings are equal to those used for segmentation stages A and B, \ie, all three keypoint prediction stages A, B, and C have an initial learning rate of $1e\mathrm{-}4$. For data augmentation, we use random horizontal mirroring only, since robustness to lighting variations is already achieved by the segmentation part, and affine transformations could easily lead to losing keypoints near the image borders, thus, invalidating the layout.

\subsection{Quantitative Results}
\label{sec:quant_results}

In the following, we quantitatively compare our method to related works and our re-implementation of RoomNet as a baseline. First, in Table~\ref{tab:lsun_quant_results}, we present results on the LSUN dataset. As can be seen, our method clearly outperforms all other methods on both error metrics, including a more advanced, recurrent version of RoomNet presented in~\cite{Lee17}.  This is in particular notable, since \cite{Mallya15, Dasgupta16, Lee17} also employ powerful semantic segmentation networks. However, these works force a single network to handle all cases, which validates our choice of dividing the task into three sub-problems.

Next, in Table~\ref{tab:hedau_quant_results}, we show results on the Hedau dataset, which already dates back to 2009. Thus, it was widely used, allowing us to give a more thorough comparison to existing approaches, including timing information (if available). Again, our method based on smart hypothesis generation is able to outperform all competing approaches, also including recent works based on Deep Learning~\cite{Mallya15, Dasgupta16, Lee17, Zou18}. In addition, it can be seen that our method is also competitive in terms of run-time, making it suitable for real-time application. In particular, it runs with approximately 12~frames per second on an NVIDIA Titan Xp GPU, which is orders of magnitude faster than most other 
approaches~\cite{DelPero12, DelPero13, Ramalingam13, Dasgupta16, Zhao17}.

Finally, in Table~\ref{tab:nyu_quant_results}, we present our performance on the NYUv2~303 dataset, where we again outperform all other RGB-based methods, and even come close to the method of Zhang~\etal~\cite{Zhang13} that additionally uses depth information.


\begin{table}[ht]
  \centering
  \begin{tabular}{lcc}
    \toprule
    Method & PE~(\%) & KPE~(\%) \\
    \midrule
    Hedau~\etal~(2009)~\cite{Hedau09} & 24.23 & 15.48 \\
    Mallya~\etal~(2015)~\cite{Mallya15} & 16.71 & 11.02 \\
    Dasgupta~\etal~(2016)~\cite{Dasgupta16} & 10.63 & 8.20 \\
    Ren~\etal~(2016)~\cite{Ren16} & 9.31 & 7.95 \\
    \textcolor[rgb]{0.5,0.5,0.5}{Zhao~\etal~(2017)~\cite{Zhao17}}\tablefootnote{Note that~\cite{Zhao17} cannot be directly compared to the other works, as it uses much richer training data that is not provided by the benchmarks.\label{fn:pio}} & \textcolor[rgb]{0.5,0.5,0.5}{5.29} & \textcolor[rgb]{0.5,0.5,0.5}{3.84} \\
    RoomNet~(rec.~3-iter.)~(2017)~\cite{Lee17} & 9.86 & 6.30 \\
    RoomNet~(re-imp.) & 11.24 & 7.14 \\
    \midrule
    Our method & \textbf{7.79} & \textbf{5.84} \\
    \bottomrule
  \end{tabular}
  \caption{Quantitative results on LSUN~\cite{Zhang16}.}
  \label{tab:lsun_quant_results}
\end{table}

\begin{table}[ht]
  \centering
  \begin{tabular}{lcc}
    \toprule
    Method & PE~(\%) & Time \\
    \midrule
    Hedau~\etal~(2009)~\cite{Hedau09} & 21.2 & - \\
    Lee~\etal~(2010)~\cite{Lee10} & 16.2 & - \\
    Wang~\etal~(2010)~\cite{Wang10} & 20.1 & - \\
    Del~Pero~\etal~(2012)~\cite{DelPero12} & 16.3 & 12~min \\
    Schwing~\etal~(2012)~\cite{Schwing12} & 12.8 & 150~ms\tablefootnote{Excluding feature computation.\label{fn:exc_feat}} \\
    Del~Pero~\etal~(2013)~\cite{DelPero13} & 12.7 & 15~min \\
    Ramalingam~\etal~(2013)~\cite{Ramalingam13} & 13.34 & 6 s\textsuperscript{\ref{fn:exc_feat}} \\
    Zhao~\etal~(2013)~\cite{Zhao13} & 14.5 & - \\
    Mallya~\etal~(2015)~\cite{Mallya15} & 12.83 & - \\
    Dasgupta~\etal~(2016)~\cite{Dasgupta16} & 9.73 & 30~s \\
    Ren~\etal~(2016)~\cite{Ren16} & 8.67 & - \\
    \textcolor[rgb]{0.5,0.5,0.5}{Zhao~\etal~(2017)~\cite{Zhao17}}\textsuperscript{\ref{fn:pio}} & \textcolor[rgb]{0.5,0.5,0.5}{6.60} & \textcolor[rgb]{0.5,0.5,0.5}{1.79~s} \\
    Zou~\etal~(2018)~\cite{Zou18} & 9.69 & 39~ms \\
    RoomNet~(rec.~3-iter.)~(2017)~\cite{Lee17} & 8.36 & 166~ms \\
    RoomNet~(re-imp.) & 12.19 & 20~ms \\
    \midrule
    Our method & \textbf{7.44} & 86~ms \\
    \bottomrule
  \end{tabular}
  \caption{Quantitative results on Hedau~\cite{Hedau09}.}
  \label{tab:hedau_quant_results}
\end{table}

\begin{table}[ht]
  \centering
  \begin{tabular}{lcc}
    \toprule
    Method & Input & PE~(\%) \\
    \midrule
    Schwing~\etal~(2012)~\cite{Schwing12} & RGB & 13.66 \\
    Zhang~\etal~(2013)~\cite{Zhang13} & RGB & 13.94 \\
    \textcolor[rgb]{0.5,0.5,0.5}{Zhang~\etal~(2013)~\cite{Zhang13}} & \textcolor[rgb]{0.5,0.5,0.5}{RGBD} & \textcolor[rgb]{0.5,0.5,0.5}{8.04} \\
    Liu~\etal~(2018)~\cite{Liu18} & RGB & 12.64 \\
    RoomNet~(re-imp.) & RGB & 12.31 \\
    \midrule
    Our method & RGB & \textbf{8.49} \\
    \bottomrule
  \end{tabular}
  \caption{Quantitative results on NYUv2~303~\cite{Zhang13}.}
  \label{tab:nyu_quant_results}
\end{table}

\renewcommand{\figWidth}{0.144}
\begin{figure*}
  \centering
  \begin{tabular}{cccccc}
    Ground Truth & Hypothesis A & Hypothesis B & Hypothesis C & Our Result & RoomNet \\
    \includegraphics[width=\figWidth\linewidth]{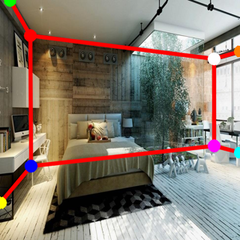} &
    \includegraphics[width=\figWidth\linewidth]{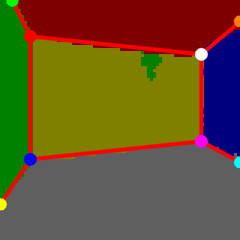} &
    \includegraphics[width=\figWidth\linewidth]{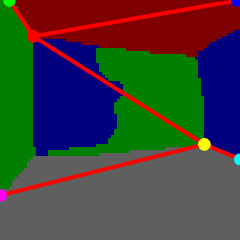} &
    \includegraphics[width=\figWidth\linewidth]{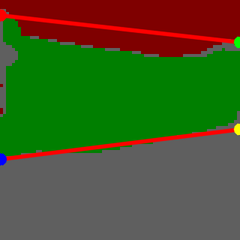} &
    \includegraphics[width=\figWidth\linewidth]{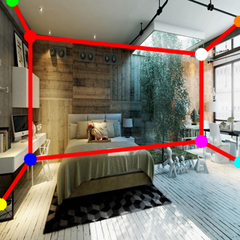} &
    \includegraphics[width=\figWidth\linewidth]{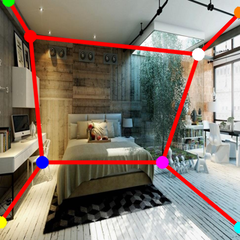} \\
    \includegraphics[width=\figWidth\linewidth]{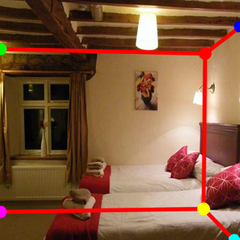} &
    \includegraphics[width=\figWidth\linewidth]{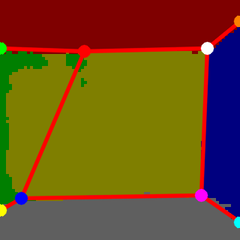} &
    \includegraphics[width=\figWidth\linewidth]{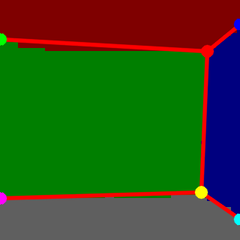} &
    \includegraphics[width=\figWidth\linewidth]{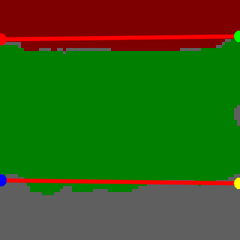} &
    \includegraphics[width=\figWidth\linewidth]{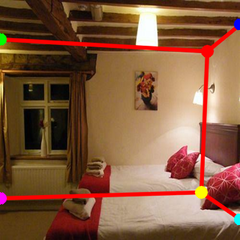} &
    \includegraphics[width=\figWidth\linewidth]{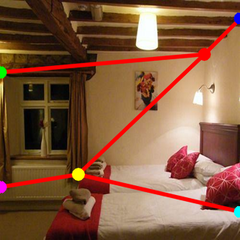} \\
    \includegraphics[width=\figWidth\linewidth]{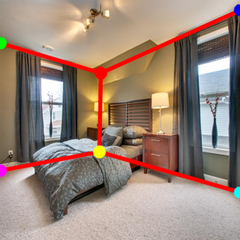} &
    \includegraphics[width=\figWidth\linewidth]{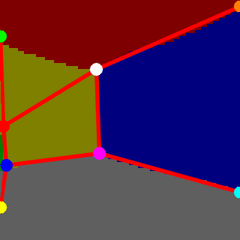} &
    \includegraphics[width=\figWidth\linewidth]{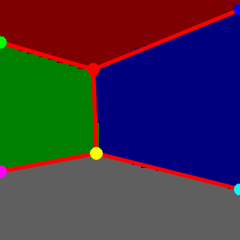} &
    \includegraphics[width=\figWidth\linewidth]{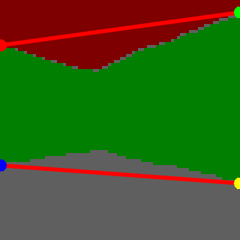} &
    \includegraphics[width=\figWidth\linewidth]{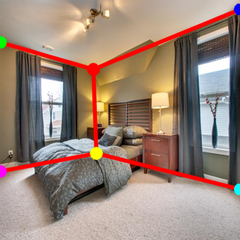} &
    \includegraphics[width=\figWidth\linewidth]{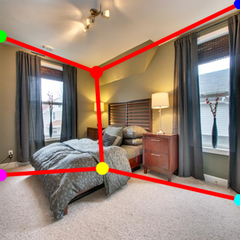} \\
    \includegraphics[width=\figWidth\linewidth]{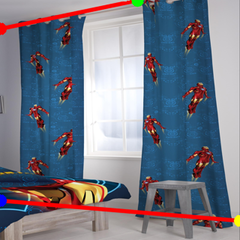} &
    \includegraphics[width=\figWidth\linewidth]{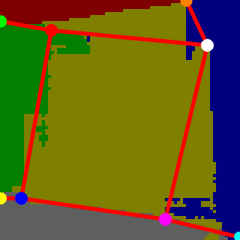} &
    \includegraphics[width=\figWidth\linewidth]{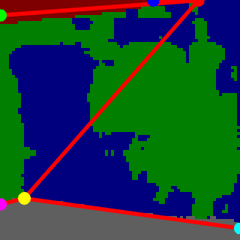} &
    \includegraphics[width=\figWidth\linewidth]{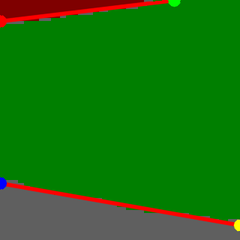} &
    \includegraphics[width=\figWidth\linewidth]{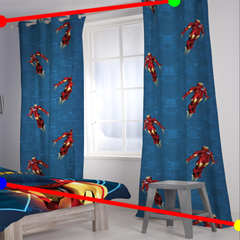} &
    \includegraphics[width=\figWidth\linewidth]{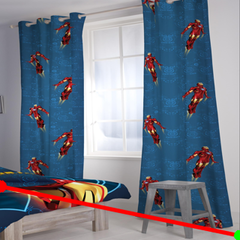} \\
    \includegraphics[width=\figWidth\linewidth]{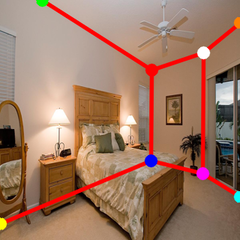} &
    \includegraphics[width=\figWidth\linewidth]{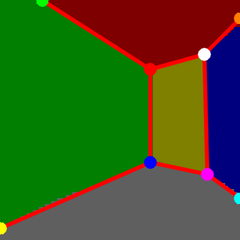} &
    \includegraphics[width=\figWidth\linewidth]{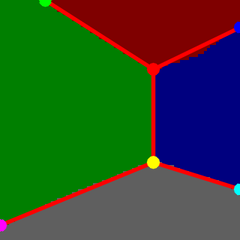} &
    \includegraphics[width=\figWidth\linewidth]{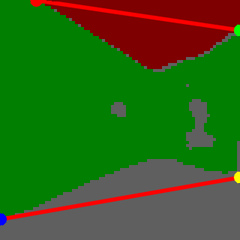} &
    \includegraphics[width=\figWidth\linewidth]{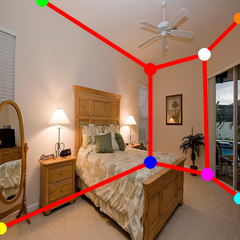} &
    \includegraphics[width=\figWidth\linewidth]{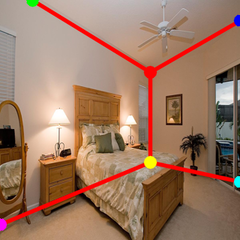} \\
    \includegraphics[width=\figWidth\linewidth]{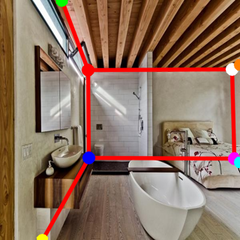} &
    \includegraphics[width=\figWidth\linewidth]{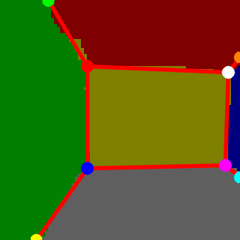} &
    \includegraphics[width=\figWidth\linewidth]{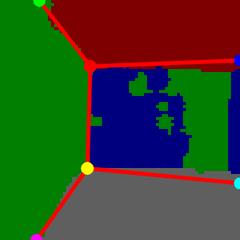} &
    \includegraphics[width=\figWidth\linewidth]{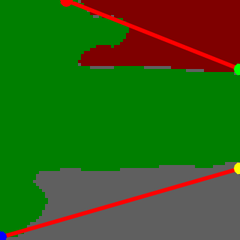} &
    \includegraphics[width=\figWidth\linewidth]{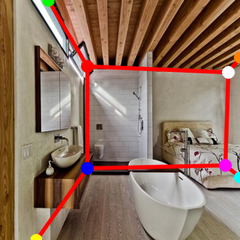} &
    \includegraphics[width=\figWidth\linewidth]{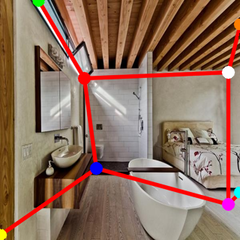} \\
    \includegraphics[width=\figWidth\linewidth]{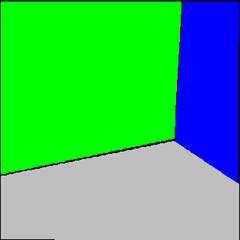} &
    \includegraphics[width=\figWidth\linewidth]{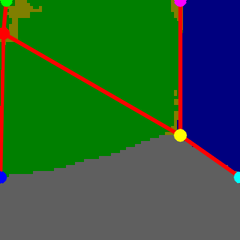} &
    \includegraphics[width=\figWidth\linewidth]{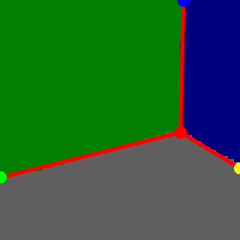} &
    \includegraphics[width=\figWidth\linewidth]{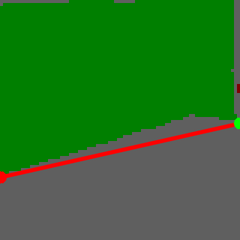} &
    \includegraphics[width=\figWidth\linewidth]{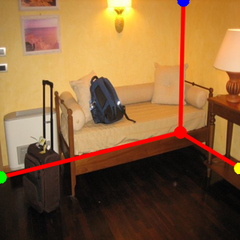} &
    \includegraphics[width=\figWidth\linewidth]{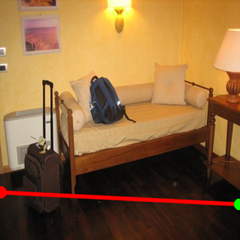} \\
    \includegraphics[width=\figWidth\linewidth]{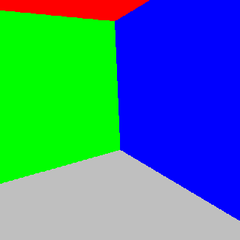} &
    \includegraphics[width=\figWidth\linewidth]{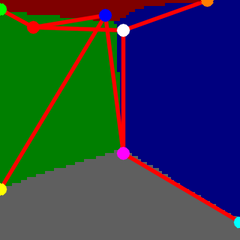} &
    \includegraphics[width=\figWidth\linewidth]{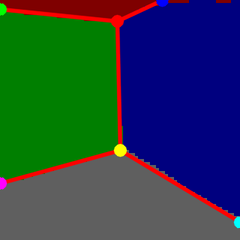} &
    \includegraphics[width=\figWidth\linewidth]{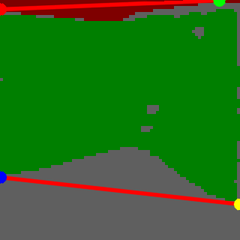} &
    \includegraphics[width=\figWidth\linewidth]{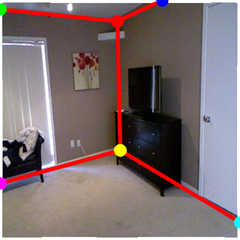} &
    \includegraphics[width=\figWidth\linewidth]{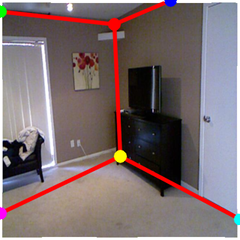} \\
  \end{tabular}
  \caption{Qualitative results showing the ground truth, our three layout hypotheses, our final result, and the result obtained with our re-implementation of RoomNet~\cite{Lee17}. Rows {1--6} present results from LSUN~\cite{Zhang16}, row 7 from Hedau~\cite{Hedau09}, and row 8 from NYUv2~303~\cite{Zhang13}. Note that the latter two do not offer ground truth keypoints, just surface labels. Best viewed in color.}
  \label{fig:qual_results}
\end{figure*}


\begin{figure*}
  \centering
  \begin{tabular}{cccccc}
    Ground Truth & Hypothesis A & Hypothesis B & Hypothesis C & Our Result & RoomNet \\
    \includegraphics[width=0.14\linewidth]{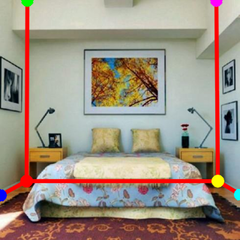} &
    \includegraphics[width=0.14\linewidth]{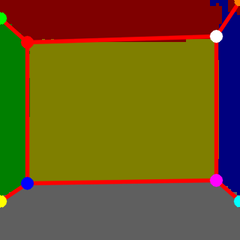} &
    \includegraphics[width=0.14\linewidth]{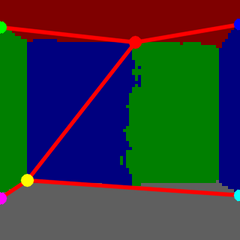} &
    \includegraphics[width=0.14\linewidth]{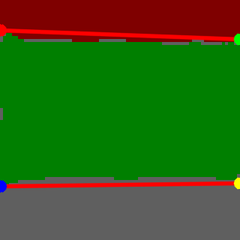} &
    \includegraphics[width=0.14\linewidth]{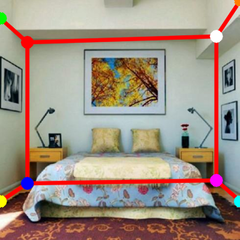} &
    \includegraphics[width=0.14\linewidth]{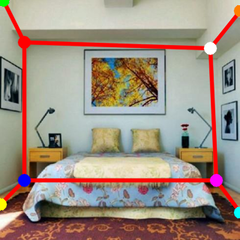} \\
    \includegraphics[width=0.14\linewidth]{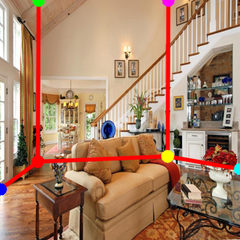} &
    \includegraphics[width=0.14\linewidth]{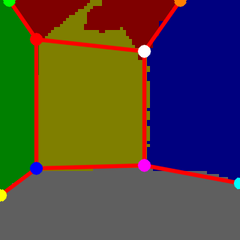} &
    \includegraphics[width=0.14\linewidth]{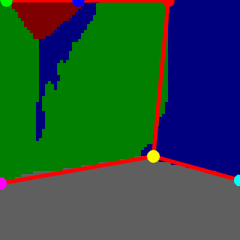} &
    \includegraphics[width=0.14\linewidth]{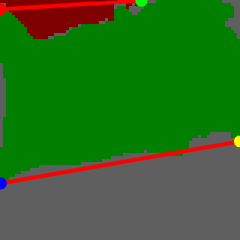} &
    \includegraphics[width=0.14\linewidth]{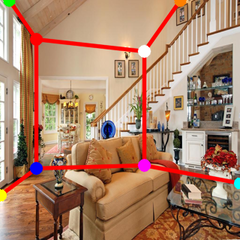} &
    \includegraphics[width=0.14\linewidth]{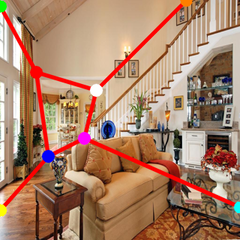} \\
  \end{tabular}
  \caption{Failure cases on LSUN~\cite{Zhang16}. The column setup is the same as in Fig.~\ref{fig:qual_results}. In the first example, the beam structures in the upper part of the image trigger a spurious ceiling region. In the second example, the cuboid layout assumption is violated. 
  }
  \label{fig:qual_failures}
\end{figure*}

\subsection{Qualitative Results}
\label{sec:qual_results}

In addition, in Fig.~\ref{fig:qual_results}, we present qualitative results generated with our method and compare it to the re-implementation of RoomNet~\cite{Lee17}. First, it is apparent that the semantic segmentation is quite robust to even severe clutter and occlusions, as can be seen from the top row for instance. Second, evaluating three specialized layout hypotheses in parallel
gives our method a clear advantage over competing approaches.
This is particularly evident in the example in the fifth row, where we can reconstruct the correct layout even though the input image provides only very little evidence.  Specifically, although hypothesis B already provides a good match, hypothesis A is even able to detect the subtle center wall in the back, giving it a higher score than B. RoomNet, on the other hand, is not able to detect this wall and generates a wrong layout estimation.  Furthermore, our method is also able to correctly predict the layout in case of the rather rare room type 6 in the fourth row, whereas the explicit type classifier of RoomNet predicts the more common type 9.

Finally, Fig.~\ref{fig:qual_failures} shows two failure cases. In the first example, the beam structures in the upper part of the image trigger a spurious, but plausibly looking ceiling region, so the predicted layout confirms it. The second example shows a room that does not follow the cuboid layout assumption, as can be seen from the tilted ceiling region in left part of the image. Nevertheless, our method is still able to provide a reasonably good, cuboid approximation.

\subsection{Segmentation as Intermediate Representation}
\label{sec:seg_representation}

To demonstrate the benefits of our intermediate representation, we perform an ablative study: Like~\cite{Lee17}, we predict the keypoints from the original image rather than from the segmentation and use the semantic segmentation only for inferring the final layout using Eq.~(\ref{eq:hypo_score}). As shown in Table~\ref{tab:ablation_results_seg}, the results clearly deteriorate across all datasets. Thus, the semantic segmentation is indeed a good intermediate representation for robustly inferring room layouts.

\begin{table}[ht]
  \centering
  \begin{tabular}{lcccc}
    \toprule
    Method & \multicolumn{2}{c}{LSUN} & Hedau & NYUv2 \\
           & PE~(\%) & KPE~(\%) & PE~(\%) & PE~(\%) \\
    \midrule
    KPs f. Img. & 12.47 & 7.36 & 11.03 & 14.33 \\
    KPs f. Seg. & \textbf{7.79} & \textbf{5.84}& \textbf{7.44} & \textbf{8.49} \\
    \bottomrule
  \end{tabular}
  \caption{Keypoint prediction from image vs. segmentation.}
  \label{tab:ablation_results_seg}
\end{table}

\subsection{Depth Estimation from Room Layouts}
\label{sec:depth_estimation}

Finally, in Fig.~\ref{fig:depth_estimation}, we show depth depth images estimated from our generated room layouts. Specifically, given a 2D layout that provides enough information (\ie, is of type~0 according to Fig.~\ref{fig:room_types}) and initializing the 3D room layout as a unit cube, we can estimate the room's height/width ratio, the focal length, and the 3D camera pose up to scale. This is achieved by iteratively minimizing the re-projection error of the four corner points at the center wall as well as additional points along the four perpendicular edges.

\renewcommand{\figWidth}{0.27}
\begin{figure}
  \centering
  \begin{tabular}{ccc}
    \includegraphics[width=\figWidth\linewidth]{sec4/lsun_01_our} &
    \includegraphics[width=\figWidth\linewidth]{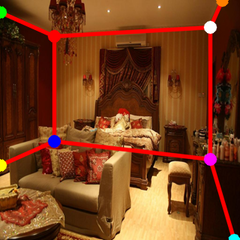} &
    \includegraphics[width=\figWidth\linewidth]{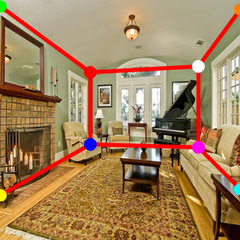} \\
    \frame{\includegraphics[width=\figWidth\linewidth]{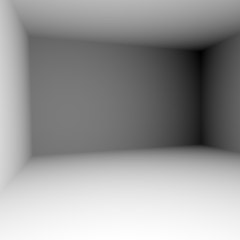}} &
    \frame{\includegraphics[width=\figWidth\linewidth]{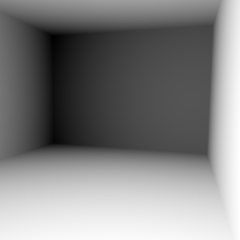}} &
    \frame{\includegraphics[width=\figWidth\linewidth]{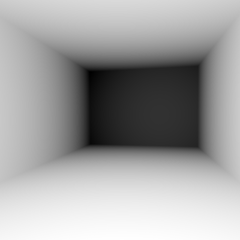}} \\
  \end{tabular}
  \caption{Estimating the relative depth (bottom row) from 2D room layouts (top row) on LSUN~\cite{Zhang16} examples.}
  \label{fig:depth_estimation}
\end{figure}


\section{Conclusion}
\label{sec:conclusion}

Estimating the layout of rooms from single images is an important but hard task. To overcome drawbacks of existing works in terms of accuracy and computational complexity, we introduce a robust and efficient hypothesize-and-test approach based on the number of visible walls. In particular, we divide the task into three sub-problems, generate a semantic segmentation and a layout hypothesis for each of them, and then select the one that has the highest consistency between these two representations. As can be seen from the experimental results, we clearly outperform the state-of-the-art on three challenging benchmark datasets,
demonstrating the benefits of our approach.
\paragraph{Acknowledgement}
This work was supported by the Christian Doppler Laboratory for Semantic 3D Computer Vision, funded in part by Qualcomm Inc.

\clearpage

%

{\small
\bibliographystyle{ieee}
\bibliography{bib_abbrevs,iccv2019}
}

\end{document}